\documentclass{article}

\PassOptionsToPackage{numbers, compress}{natbib}
\usepackage[preprint]{neurips_2022}

\usepackage[utf8]{inputenc} 
\usepackage[T1]{fontenc}    
\usepackage{hyperref}       
\usepackage{url}            
\usepackage{booktabs}       
\usepackage{amsfonts}       
\usepackage{nicefrac}       
\usepackage{microtype}      
\usepackage{xcolor}         

\usepackage{graphicx}
\usepackage{amsmath,amssymb}
\usepackage{color}
\usepackage[ruled,linesnumbered]{algorithm2e}
\usepackage{array}
\usepackage{url} 
\usepackage{amsfonts}

\usepackage{pifont}

\usepackage{xpatch}

\usepackage{wrapfig}

\title{Understanding Weight Similarity of Neural Networks via Chain Normalization Rule and Hypothesis-Training-Testing}

\author{%
  Guangcong Wang\\
  Sun Yat-sen University\\
  wanggc3@mail2.sysu.edu.cn
  \And
  Guangrun Wang\\
  University of Oxford\\
  wanggrun@gmail.com\\
  \And
  Wenqi Liang\\
  Sun Yat-sen University\\
  wenqi.liang.sysu@gmail.com
  \And
  Jianhuang Lai\thanks{Corresponding Author}\\
  Sun Yat-sen University\\
  stsljh@mail.sysu.edu.cn\\
}

\begin{document}

\maketitle

\begin{abstract}
We present a weight similarity measure method that can quantify the weight similarity of non-convex neural networks. To understand the weight similarity of different trained models, we propose to extract the feature representation from the weights of neural networks. We first normalize the weights of neural networks by introducing a chain normalization rule, which is used for weight representation learning and weight similarity measure. We extend the traditional hypothesis-testing method to a hypothesis-training-testing statistical inference method to validate the hypothesis on the weight similarity of neural networks. With the chain normalization rule and the new statistical inference, we study the weight similarity measure on Multi-Layer Perceptron (MLP), Convolutional Neural Network (CNN), and Recurrent Neural Network (RNN), and find that the weights of an identical neural network optimized with the Stochastic Gradient Descent (SGD) algorithm converge to a similar local solution in a metric space. The weight similarity measure provides more insight into the local solutions of neural networks. Experiments on several datasets consistently validate the hypothesis of weight similarity measure.
\end{abstract}

Neural networks have achieved remarkable empirical success in a wide range of machine learning tasks \cite{lecun1989backpropagation,krizhevsky2012imagenet,he2016deep} by finding a good local solution. How to better understand the characteristics of local solutions of neural networks remains an open problem. Lots of evidence shows that the local solutions of an identical neural network optimized by SGD achieve nearly the same classification accuracy. Are these local solutions (trained weights) equivalent? Existing methods \cite{li2016convergent,raghu2017svcca,morcos2018insights,kornblith2019similarity} mainly focus on analyzing the input-output activation of neural networks while ignoring the weight analysis of neural networks. A question arises: are the weights of trained neural networks similar in different runs?  

In this paper, we make the first attempt to measure the weight similarity of neural networks and provide a new avenue to the understanding of local solutions of neural networks. Let $f(\mathbf{x};W)$ be a neural network. We expect the weight $W$ can be transformed into a ``common data point". In this way, we can learn to measure weight similarity by conventional data similarity approaches. To achieve this goal, we propose a chain normalization rule to normalize hierarchical weight $W$. The normalized weight can be further used as ``common" data for the similarity learning and measure. To validate the weight similarity, we add a training technique to hypothesis-testing and develop a hypothesis-training-testing method to solve a general statistical inference. With the chain normalization rule and the new statistical inference method, we can better understand the local solutions of neural networks.

Naturally, weight similarity measure is similar to data similarity measure. Given a set of data points $\{{\mathbf{x}}^i\}_{i=1}^K$, data similarity measure is to find a similarity measure function $f(\cdot;W)$ that quantifies the data similarity. In a similar way, given a set of weights $\{W^{i}\}_{i=1}^K$ (each of which is the weight of a neural network), weight similarity measure is to find a similarity measure function $F(\cdot;\Theta)$ that describes the weight similarity between $\{W^{i}\}_{i=1}^K$. In the data metric space, data similarity measure can uncover the hidden relationship from big data. In the weight metric space, weight similarity measure can uncover the underlying relationship of local solutions of neural networks. One of the key differences is that the weights of neural networks are hierarchical and randomly permuted, which differ from traditional data points.

Overall, we mainly make several contributions as follows.
\begin{itemize}
  \item We propose a weight similarity measure method to quantify the weight similarity of an identical neural network optimized with SGD. We propose a chain normalization rule to normalize the randomly permuted weights of neural networks.
  \item We extend the traditional hypothesis-testing statistical inference to a hypothesis-training-testing method with the addition of a training technique, which can solve general statistical inference problems. We apply it to validate the hypothesis regarding the weight similarity of neural networks. We find that the local solutions (weights) optimized by the Stochastic Gradient Descent (SGD) algorithm are similar in a metric space.
  \item We study the weight similarity measure of Multi-Layer Perceptron (MLP), Convolutional Neural Network (CNN), and Recurrent Neural Network (RNN) on the tiny ImageNet, CIFAR-100, and NameData datasets, respectively. The experiments show consistent testing results for the hypothesis.
  \item We analyze several factors of neural networks and find that 1) adding one layer after the last layer or changing the ReLU function into the leaky ReLU has little impact on local solutions; 2) changing plain networks into residual networks has some impact on local solutions.
 \end{itemize}

\section{Related Work}
\label{sec:related}
Neural networks are often regarded as black boxes due to the non-convexity. To better understand these black boxes, various approaches provide effective tools for visual interpretability of neural networks \cite{simonyan2013deep,dosovitskiy2016inverting,zeiler2014visualizing,Zhou2015Learning,Selvaraju2016Grad}. These approaches apply the gradient of the class scores with respect to the input or exploited the de-convolution operations to visualize the attention activation at high-level semantics. Recent studies \cite{raghu2017svcca,morcos2018insights,kornblith2019similarity} focused on studying the similarity of activation representation at the output of neural networks, which fails to directly uncover the internal information of network networks. In this paper, we open the black box and study the weight similarity of internal structures.

Our proposed method is also inspired by meta-learning. Meta-learning \cite{zoph2016neural,andrychowicz2016learning,finn2017model,liu2018progressive} focuses on automated learning algorithms by using learning to learn models to search hyper-parameters. Basically, they aim to learn hyper-parameters to learn weights of neural networks. For example, Zoph and Le \cite{zoph2016neural} used a recurrent network to generate the model descriptions of neural networks and trained the RNN with reinforcement learning to maximize the expected accuracy of the generated architectures on a validation set. Andrychowicz et al. \cite{andrychowicz2016learning} attempted to learn an LSTM-based neural optimizer to learn how to optimize neural networks. Finn et al. \cite{finn2017model} introduced a meta-learning method based on learning easily adaptable model parameters through gradient descent. These methods learn hyper-parameters to automatically learn weights of neural networks for data representations. Our proposed method regards weight of networks as ``new data", and directly learns weight representations from the weights of neural networks.

\section{Hypothesis-Training-Testing}
\label{sec:htt}

We first introduce a new statistical inference, which is used for weight similarity in the next section. We present a brief review of the traditional hypothesis-testing. We then propose an improved version of the hypothesis-testing termed hypothesis-training-testing.




\textbf{Hypothesis-Testing} Generally, a hypothesis test begins with an assumption about the statistical relationship. We formulate this assumption as a null hypothesis $\mathcal{H}_0$ and design an alternative hypothesis $\mathcal{H}_1$, where $\mathcal{H}_0$ and $\mathcal{H}_1$ are mutually exclusive. We pre-define a significance level of $\alpha$. Based on $\mathcal{H}_0$ and $\mathcal{H}_1$, we directly calculate the test statistic and obtain a $p$-value. If the $p$-value lies within the confidence level (determined by $\alpha$), we accept the null hypothesis $\mathcal{H}_0$ and reject $\mathcal{H}_1$, or we should reject the null hypothesis $\mathcal{H}_0$ and accept $\mathcal{H}_1$.


The key point of hypothesis-testing is to compare the test statistic with the hypothesis statistic. However, traditional hypothesis-testing is limited to special probability distributions (e.g., normal distribution) and statistical characteristics (e.g., mean value and variance). In machine learning,  we often need to consider a general case. Therefore, we generalize probability distributions to any function $F$ and generalize statistical characteristics to any parameter $\Theta$. The goal is to estimate the parameter $\Theta$ and compare it with the hypothesis parameter $\Theta_0$. In this case, it is difficult for traditional hypothesis-testing to directly calculate ${\Theta}$ because ${\Theta}$ is not a statistical characteristic. To achieve it, we add a training approach to hypothesis-testing to learn the parameter ${\Theta}$.

\textbf{Hypothesis-Training-Testing} To solve the generalized test problem, we extend the hypothesis-testing to the hypothesis-training-testing with the addition of a training process, which can learn the parameter $\Theta$ according to the hypothesis. The hypothesis-training-testing consists of three main steps. \textbf{First}, let $\mathbf{x}_1$, $\mathbf{x}_2$, ..., $\mathbf{x}_K$ be sample data and $F(\cdot;\Theta)$ be a function parameterized by $\Theta$. Similar to the hypothesis-testing, we make a null hypothesis $\mathcal{H}_0$ and an alternative hypothesis $\mathcal{H}_1$ about the population data. \textbf{Second}, based on $\mathcal{H}_0$ and $\mathcal{H}_1$, we construct a set of assumptive labels $\{y_i\}_{i=1}^K$ and assign them for the data set, i.e., $(\mathbf{x}_1,y_1)$, $(\mathbf{x}_2,y_2)$, ..., $(\mathbf{x}_K,y_K)$. Unlike hypothesis-testing, the sample data are split into a training data set $D_{train}$ and a test data set $D_{test}$ without overlapping, where $D_{train}=\{(\mathbf{x}_1,y_1),(\mathbf{x}_2,y_2),...,(\mathbf{x}_Q,y_Q)\}$ and $D_{test}=\{(\mathbf{x}_{Q+1},y_{Q+1}),(\mathbf{x}_{Q+2},y_{Q+2}),...,(\mathbf{x}_{K},y_{K})\}$. We train a model $F(\cdot;\Theta)$ using the training data set $D_{train}$. \textbf{Third}, with the learned $\Theta$, we test the error on  $D_{test}$. If the test error on $D_{test}$ falls below a pre-defined significance level of $\alpha$, we accept the null hypothesis $\mathcal{H}_0$ and reject $\mathcal{H}_1$, and vice versa.

There are several differences between hypothesis-testing and hypothesis-training-testing. \textbf{First}, hypothesis-testing directly calculates the test statistic $\Theta$ according to the hypotheses. Hypothesis-training-testing introduces a training approach to learn the parameter $\Theta$ using the assumptive labels. The assumptive labels are based on the hypotheses. \textbf{Second}, hypothesis-testing is limited to special probability distributions and statistical characteristics, whereas hypothesis-training-testing does not have such a limitation. \textbf{Third}, hypothesis-training-testing split the data set into a training set and a test set, whereas hypothesis-testing does not. In the next section, we will discuss how to use hypothesis-training-testing to verify the weight similarity hypothesis of neural networks.

\section{Weight Similarity Measure of Neural Networks}
\label{sec:ssm}
\subsection{Notation and Problem Definition}
Let $f(\cdot;W)$ be a neural network that contains a sequence of stacked units $\{u_l\}_{l=1}^{L}$. A unit $u_l$ contains a linear operator $g_l(\cdot;W_l)$ and an activation function $\sigma(\cdot)$. We formulate the neural network as:\begin{equation}
    \label{eq:net}
f(\mathbf{x};W)=\sigma (g_L(\sigma (g_{L-1}(...\sigma (g_1(\mathbf{x};W_1))...;W_{L-1}));W_L)),
\end{equation}where $n_{l}$ denotes the number of neurons at the $l$-th layer, $\mathbf{x} \in \mathbb{R}^{n_{0}}$ is the input, and $W_l \in \mathbb{R}^{n_{l-1}\times n_{l}}$ denotes the weight of the $l$-th linear transformation. $W=(W_1,W_2,...,W_{L-1},W_L)$ denotes the weight of the neural network. We denote $h_l = \sigma (g_{l}(...\sigma (g_1(x;W_1))...;W_{l}))\in \mathbb{R}^{n_l}$ as the output of the $l$-th hidden layer. Because a CNN can be regarded as a patch-based MLP and an RNN can be regarded as a cyclic version of MLP, we use MLP for discussion without loss of generality. Given a learning task $\mathcal{T}$ over a dataset $\mathcal{D}$, we train a neural network $f(\cdot;W)$ with SGD and obtain a local solution $W$. We repeat this training process $N$ times and obtain $N$ local solutions $\{W^{\mathcal{T},i}\}_{i=1}^N$. The goal of weight similarity measure of neural networks is to study the similarity between $\{W^{\mathcal{T},i}\}_{i=1}^N$.

\subsection{Hypothesis-Training-Testing on Solution Similarity of Neural Networks}
\label{subsec:htt-nn}
We use the hypothesis-training-testing statistical inference to solve weight similarity measure, which consists of three main steps: hypothesis, training, and testing.

\textbf{Hypothesis} To measure the weight similarity of neural networks, we begin the hypothesis-training-testing statistical inference by making the null hypothesis $\mathcal{H}_0$, which is given as follows:\begin{equation}
    \label{eq:h_0}
\mathcal{H}_0:{W^{\mathcal{T},1}\triangleq W^{\mathcal{T},2}\triangleq ...\triangleq W^{\mathcal{T},N}},
\end{equation}where Eq. \eqref{eq:h_0} assumes that a set of local solutions of an identical network in the learning task $\mathcal{T}$ are equivalent in a metric space. When the objective is strongly convex, Eq. \eqref{eq:h_0} holds because the convex objective produces a unique global minimum; that is, every training process produces the same solution. When the objective is non-convex, it leads to a set of local solutions,  and it is NP-hard to find the global minimum. Intuitively, non-convex problems have different local minima, which often leads to significantly different classification/regression accuracies in machine learning. However, it is widely observed that the optimization of neural networks can produce a similar accuracy by the SGD optimizer when the training process is repeated. Therefore, we make an intuitive hypothesis as in Eq. \eqref{eq:h_0}. Next, we develop a weight similarity measure to explore the underlying relationship between local solutions.

\textbf{Training} In Eq. \eqref{eq:h_0}, we assume that the weights $\{W^{\mathcal{T},i}\}_{i=1}^N$ share an equivalent solution (a class). To verify this hypothesis, we regard $W^{\mathcal{T},i}$ as a ``data" point and design a weight representation for similarity measure. To learn a weight representation, we collect $M$ learning tasks, which can be thought as analogous to $M$ classes. We denote these learning tasks as $\Sigma=\{\mathcal{T}_1,\mathcal{T}_2,...,\mathcal{T}_M\}$. For any learning task $\mathcal{T}_j\in \Sigma~(1\leqslant j \leqslant M)$, we generate $N_j$ local solutions, denoted as $\{W^{\mathcal{T}_j,i}\}_{i=1}^{N_j}$. Hence, we obtain a weight set $D=\{\{W^{\mathcal{T}_1,i}\}_{i=1}^{N_1},\{W^{\mathcal{T}_2,i}\}_{i=1}^{N_2},...,\{W^{\mathcal{T}_M,i}\}_{i=1}^{N_M}\}$. As assumed in Eq. \eqref{eq:h_0}, for any $j\in [1,M]$, $\{W^{\mathcal{T}_j,i}\}_{i=1}^{N_j}$ share a similar weight, which can be thought as analogous to an object class. We assign them the weight label $\mathcal{Y}_{\mathcal{T}_j}$. Therefore, $M$ learning tasks can be thought as analogous to $M$ object classes. We train a weight classifier $F$ to distinguish weight classes, given by:\begin{equation}
    \label{eq:train}
\mathcal{L}=\sum\limits_{j=1}^M\sum\limits_{i=1}^{N_j}\ell(F(W^{\mathcal{T}_j,i};\Theta),\mathcal{Y}_{\mathcal{T}_j}),
\end{equation}where $\mathcal{Y}_{\mathcal{T}_j}$ denotes the weight class generated from a learning task $\mathcal{T}_j$. $F(W^{\mathcal{T}_j,i};\Theta)$ denotes a predictor parameterized by $\Theta$, which maps a weight $W^{\mathcal{T}_j,i}$ to a weight label. $l$ denotes the prediction loss. $\mathcal{L}$ denotes the empirical loss over the entire weight set $D$. Because a weight $W^{\mathcal{T}_j,i}$ contains weights of $L$ layers of neural networks, we normalize the weights before feeding them into a weight classifier. We re-write Eq. \eqref{eq:train} as:\begin{equation}
    \label{eq:train2}
\mathcal{L}=\sum\limits_{j=1}^M\sum\limits_{i=1}^{N_j}\ell(F(\phi(W^{\mathcal{T}_j,i});\Theta),\mathcal{Y}_{\mathcal{T}_j}),
\end{equation}where $\phi(\cdot)$ denotes a weight normalization operator of a neural network, which transforms the weight $W^{\mathcal{T}_j,i}$ into ``common data". In this way, we can choose a traditional data classifier for weight similarity learning. Therefore, the most important point is to design a good $\phi(\cdot)$. We leave the discussion of $\phi(\cdot)$ in Section \ref{topo} and analyze $F(\cdot;\Theta)$ in Section \ref{subsec:learn}.

\textbf{Testing} After obtaining a trained weight classifier, we test the hypothesis $\mathcal{H}_0$ in Eq. \eqref{eq:h_0}. To do this, we test the accuracy of the classifier $F(\phi(\cdot);\Theta)$ on a test set. We use two evaluation protocols to test $\mathcal{H}_0$ following traditional data similarity learning.

In the first evaluation protocol, we adopt weight classification which is similar to image classification. For any learning task $\mathcal{T}_j\in \Sigma~(1\leqslant j \leqslant M)$, we generate another $N_j^{'}$ local solutions by repeating the training of the neural network. The $N_j^{'}$ weights of $\mathcal{T}_j$ differ from the $N_j$ ones of $\mathcal{T}_j$ in the training set. Therefore, we obtain a test set including $N_1^{'}+N_2^{'}+...+N_M^{'}$ local solutions. Only if $F(\phi(\cdot);\Theta)$ can capture the characteristics of the weights of $\mathcal{T}_j$, can $F(\phi(\cdot);\Theta)$ provide an accurate prediction for the new weights in the test set. This evaluation protocol quantifies the capacity of a weight classifier to describe the characteristics of weights. If a weight classifier obtains an error on the test set lower than a pre-defined significance level $\alpha$, we accept the hypothesis $\mathcal{H}_0$ in Eq. \eqref{eq:h_0}, and vice versa.

In the second evaluation protocol, we adopt weight retrieval which is similar to image retrieval. We use the last fully-connected layer of $F(\phi(\cdot);\Theta)$ as the weight representation for weight retrieval. For weight retrieval, we generate another $M^{'}$ unseen learning tasks for testing, which are denoted as $\Sigma^{'}=\{\mathcal{T}_{M+1},\mathcal{T}_{M+1},...,\mathcal{T}_{M+M^{'}}\}$, where $\Sigma^{'}\bigcap\Sigma=\emptyset$. For any learning task $\mathcal{T}_j~(M+1\leqslant j\leqslant M+M^{'})$, we generate $N_j$ local solutions, among which we select some weights as queries and the other weights for the gallery. Given a query weight, the goal of retrieval is to search the weights that share the same weight label with the query from the gallery. Not only can this evaluation protocol validate the hypotheses, but also it can measure the generalization of the weight representation for the unseen learning tasks. If a trained weight classifier obtains a retrieval error lower than a pre-defined significance level of $\alpha$ in the test set, we accept the hypothesis $\mathcal{H}_0$, and vice versa.



\begin{figure*}
 \vspace{-44pt}
  \centering
  \includegraphics[width=0.85\textwidth]{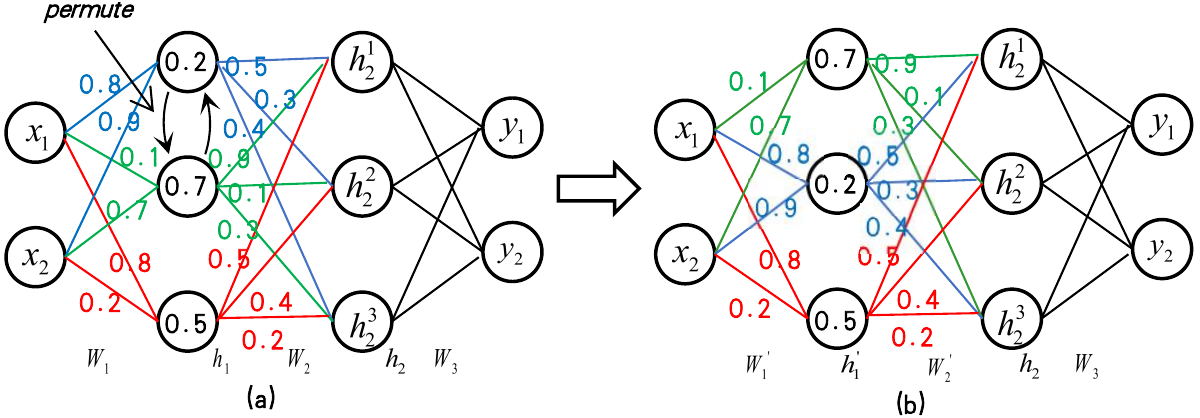}
  \vspace{-9pt}
  \caption{\small{Equivalent topology of two neural networks. \textbf{(a)} The network contains two intermediate hidden layers, denoted $\mathbf{h}_1=(0.2,0.7,0.5)^T$ and $\mathbf{h}_2=(h_2^1,h_2^2,h_2^3)^T$. The weights $W_1$ and $W_2$ are $[0.8,0.9;0.1,0.7;0.8,0.2]^{T}$ and $[0.5,0.9,0.5;0.3,0.1,0.4;0.4,0.3,0.2]^{T}$ (MATLAB-like notation), respectively. \textbf{(b)} We permute the first and second neurons of $\mathbf{h}_1$ and permute the corresponding columns of $W_1$ and rows of $W_2$. We obtain $\mathbf{h}_1^{'}=(0.7,0.2,0.5)^T$ and $W_1^{'}=[0.1,0.7;0.8,0.9;0.8,0.2]^{T}$ and $W_2^{'}=[0.9,0.5,0.5;0.1,0.3,0.4;0.3,0.4,0.2]^{T}$. Given any input $\mathbf{x}$, the output of these two functions (a) and (b) are equivalent because $W_2^TW_1^T\mathbf{x}=(W_2^{'})^T(W_1^{'})^T\mathbf{x}$.}}\label{fig:problem}
  \vspace{-14pt}
\end{figure*}

\subsection{Weight Normalization of neural networks}
\label{topo}
In Section \ref{sec:ssm}, we train a weight classifier by feeding weights and their labels. To transform the weight similarity measure problem into the data similarity measure problem, we have to transform the weights into ``common" data. In Eq. \eqref{eq:train2}, we simply use the notation $\phi(\cdot)$ for the weight normalization of neural networks. In this section, we present the details of weight normalization operator $\phi(\cdot)$. Specifically, in Section \ref{subsec:random}, we first study the permutation of neural networks. We then propose a chain normalization rule for weight normalization $\phi(\cdot)$ in Section \ref{subsec:chain}.

\subsubsection{Permutation of Neural Networks}
\label{subsec:random}
To transform the weights of neural networks as ``common data," we study the permutation of neurons and weights for weight normalization. At the $1$-st layer and the $L$-th layer, permutation of neurons are commonly pre-defined by the structure of the data and label, respectively. At the $l$-th intermediate layer ($2\leqslant l\leqslant L-1$), the permutation of neurons are constrained by the $(l-1)$-th layer and the $(l+1)$-th layer. Given an input, we analyze the permutation of intermediate neurons and weights but keep the output unchanged. Let us consider a 4-layer linear MLP for simplification, as shown in \textbf{Fig. \ref{fig:problem}}. Given any input $\mathbf{x}$, the output of these two network functions (a) and (b) are equivalent because $W_2^TW_1^T\mathbf{x}=(W_2^{'})^T(W_1^{'})^T\mathbf{x}$. Weight permutation is exchanged in accord with neuron permutation  (e.g., blue and green connections are exchanged). Due to neuron permutation and weight permutation, repeating the optimization procedures of an identical neural network generates different permutations of equivalent local solutions. Next, we discuss the weight normalization of neural networks for weight representation and similarity measure.

\subsubsection{Chain Normalization Rule of Neural Networks}
\label{subsec:chain}
One intuitive approach to normalize the weights of neural networks is to transform $W_1$, $W_2$, ... , $W_{L-1}$, $W_L$ into the normalized forms $W_1^{*}$, $W_2^{*}$, ... , $W_{L-1}^{*}$, $W_L^{*}$ that are invariant to the weight permutation. Due to random weight permutation in each training process, we must eliminate the permutation variables such that the weight similarity measure is invariant to the permutation of weights. We begin by considering the weight $W_1\in \mathbb{R}^{n_0\times n_1}$. If we permute the $i$-th and $j$-th neurons of $h_1$ and wish to keep the output unchanged, we have to permute the $i$-th and $j$-th columns of $W_1$ and the $i$-th and $j$-th rows of $W_2$ correspondingly, as illustrated in \textbf{Fig. \ref{fig:problem}}. Let $Q_1\in \mathbb{R}^{n_1\times n_1}$ be a column permutation matrix (i.e., a matrix formed by permuting the columns of the identity matrix $I\in \mathbb{R}^{n_1\times n_1}$) and $W_1^{*}$ be the normalized form of $W_1$. The weight normalization of $W_1$ is given by $W_1Q_1=W_1^{*}$. However, we cannot directly obtain $W_1^{*}$, because both $W_1^{*}$ and $Q_1$ are unknown. In fact, permutation of $W_1$ changes in different runs, so $Q_1$ is not fixed either. We compute $(W_1Q_1)(W_1Q_1)^{T}=W_1^{*}W_1^{*T}$ by multiply the transpose for both sides. Since $Q_1$ is the permutation of the identity matrix $I$, $Q_1Q_1^{T}=I^{n_1\times n_1}$, we obtain:\begin{equation}
\label{eq:tranform2_2}
W_1W_1^{T}=W_1^{*}W_1^{*T}.
\end{equation}Since $W_1^{*}W_1^{*T}$ is a normalized value and is fixed, $W_1W_1^{T}$ is invariant to the random permutation $Q_1$. We can generalize to $W_l$, which is given by:\begin{equation}
\label{eq:tranform7_}
W_1W_2...W_lW_l^T...W_2^TW_1^T=W_1^{*}W_2^{*}...W_l^{*}W_l^{*T}...W_2^{T*}W_1^{T*}.
\end{equation}The detailed proof is provided in \textbf{Appendix}. It is observed that the left side of Eq. \eqref{eq:tranform7_} is independent of permutation factors because the right side comprises a sequence of normalized weights and is thus fixed. We term Eq. \eqref{eq:tranform7_} as the \emph{chain normalization rule} for weight normalization of a neural network. Here, a $l$-th chain is defined as a sequence of layers that begin with the first layer to the $l$-th layer. The chain normalization rule suggests that we can normalize the weights of neural networks based on chains instead of layers. Therefore, the weight normalization operator $\phi(\cdot)$ in Eq. \eqref{eq:train2} is given by:
\begin{equation}
\label{eq:tranform8}
\phi(W_1,W_2,...,W_l)=W_1W_2...W_lW_l^T...W_2^TW_1^T,
\end{equation}
where $\phi(\cdot)$ takes a chain (from $W_1$ to $W_l$) as the input and produces a matrix as a normalized weight. In this way, we transform the layer-based weights $W_1$, $W_2$, ... , $W_L$ into the chain-based weights $\phi(W_1)$, $\phi(W_1, W_2)$, ... , $\phi(W_1, W_2,...,W_L)$, or briefly denoted as $\phi^1(W)$, $\phi^2(W)$, ... , $\phi^L(W)$ for simplification. The weight of a neural network with $L$ layers is split into $L$ chains for normalization. 


\subsection{Weight Similarity Measure}
\label{subsec:learn}
In Section \ref{subsec:chain}, we transform the weights as ``common" data points. The solution similarity measure problem is transformed into the conventional data similarity measure problem. We use cross-entropy loss for similarity learning. For weight classification, we directly use the maximum classification scores. For weight retrieval, we use the last fully-connected layers for weight representation. We train $L$ weight classifiers for $L$ types of chains. For the $l$-th weight classifier $F_l$, we re-write Eq. \eqref{eq:train2} as:\begin{equation}
    \label{eq:train3}
\mathcal{L}_l=\sum\limits_{j=1}^M\sum\limits_{i=1}^{N_j}\ell(F_l(\phi^l(W^{\mathcal{T}_j,i});\Theta_l),\mathcal{Y}_{\mathcal{T}_j}),
\end{equation}where $1\leqslant l \leqslant L$. In this paper, we adopt a simple weight classifier $F_l$ that includes three components: 1) $F_l$ takes $\phi^l(W^{\mathcal{T}_j,i})\in \mathbb{R}^{n_0\times n_0}$ as the input and reshapes it into a vector $\mathcal{X}^l\in \mathbb{R}^{n_0^2}$; 2) $F_l$ then projects $\mathcal{X}^l$ into a common weight representation space by learning a projection matrix $\Theta_l\in \mathbb{R}^{n_0^2\times M}$, i.e., $\Theta_l^T\mathcal{X}^l$; 3) $F_l$ normalizes the projected weight representation by using a softmax function (denoted $soft(\cdot)$) and outputs a probability distribution vector $q^l\in \mathbb{R}^{M}$, i.e., $q^l=soft(\Theta_l^T\mathcal{X}^l)$; The conventional cross-entropy loss for the weight classifier is given by:\begin{equation}
    \label{eq:loss}
    \ell(q^l,\mathcal{Y}_{\mathcal{T}_j}) =  - \sum\limits_{k = 1}^M {{{\mathcal{Y}_k}}\log({q_k^l})},
\end{equation}where ${{\mathcal{Y}_k}}$ is the $k$-dimensional value of the one-hot label of $\mathcal{Y}_{\mathcal{T}_j}$. The $q_k^l$ represents the probability of the $k$-th weight class of the $l$-th chain.

\section{Experiment}
\label{sec:exp}

\subsection{Hypothesis-training-testing on tiny ImageNet}
\label{subsec:tiny}

The tiny ImageNet dataset, which is drawn from the ImageNet \cite{russakovsky2015imagenet}, has 200 classes, $64\times 64$ in size. On this dataset, we evaluate the weight similarity measure on MLP and CNN, respectively.  

\textbf{Hypothesis} The goal of this experiment is to validate the null hypothesis $\mathcal{H}_0$ that the SGD-based local solutions of a non-convex neural network are similar, as discussed in Eq. \eqref{eq:h_0}.

\textbf{Training} We first generate the weight set $\{W^{\mathcal{T}_j,i}\}_{i=1}^{N_j}$ based on the null hypothesis $\mathcal{H}_0$. We train a 5-layer convolutional network (PlainNet-5) and a 4-layer MLP (MLP-4) to create two weight sets. We divide 200 image classes of the tiny ImageNet dataset into 50 groups as 50 data subsets. Each data subset contains 4 image classes and represents a learning task. Therefore, we have 50 learning tasks whose weight labels range from $0$ to $49$. For each learning task,  we repeat the training procedure 100 times to obtain 100 local solutions (weights). Therefore, we generate 5,000 weights for MLP-4 and PlainNet-5, respectively. We use SGD as the optimizer. After generating the weight sets, we then train weight classifiers for CNN and MLP, respectively.

\begin{figure*}
  \vspace{-46pt}
  \centering
  \includegraphics[width=1.0\textwidth]{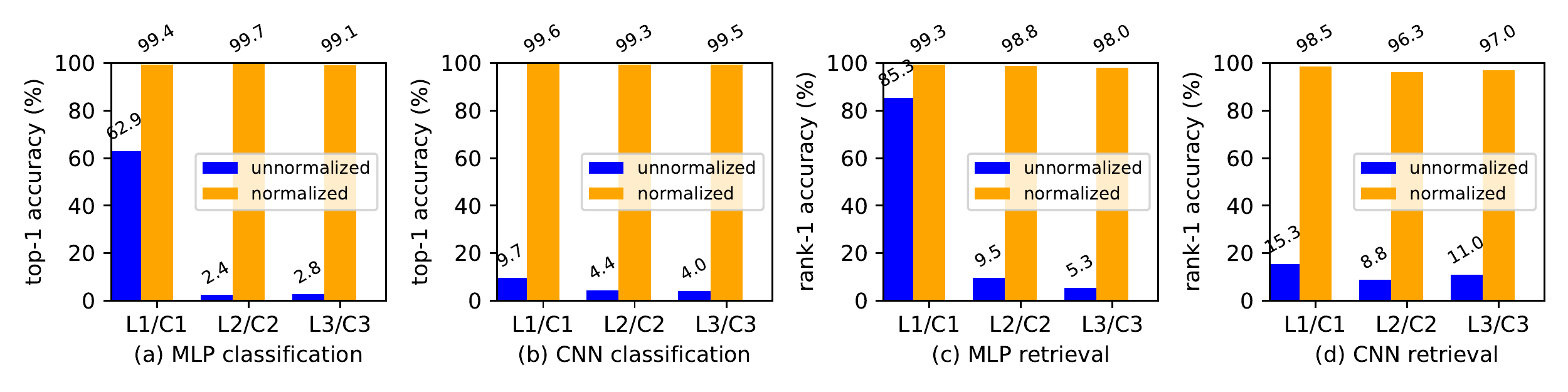}
  \vspace{-22pt}
  \caption{\small{Hypothesis-training-testing on the tiny ImageNet Dataset. L1, L2, and L3 denote the first, second, and third layers and are the x-axis of the unnormalized method. C1, C2, and C3 denote the first, second, and third chains and are the x-axis of the normalized method. In (a) and (b), we adopt the weight classification evaluation protocol for MLP and CNN, respectively. In (c) and (d), we adopt the weight retrieval evaluation protocol for MLP and CNN, respectively.}}\label{fig:tiny}
  \vspace{-11pt}
\end{figure*}

\textbf{Testing} We adopt two kinds of evaluation protocols for hypothesis testing, as discussed in Section \ref{sec:ssm}. We use a significance level of $\alpha=5\%$. For weight classification, we evaluate the accuracy of weight classification for MLP and CNN, respectively. We predict the labels of weights by the chain normalization rule and the weight classifier $F(\cdot;\Theta)$ in Eq. \eqref{eq:train3}. For each weight class, we sample 60 weights for training and the other 40 for testing. The training set contains 3,000 weights, and the test set contains 2,000. We train the weight representation by classifying 50 weight classes. The experimental results are shown in \textbf{Fig. \ref{fig:tiny} (a)} and \textbf{(b)}. In weight classification, the testing achieves 99.4\%, 99.7\%, and 99.1\% top-1 accuracy for MLP and 99.6\%, 99.3\%, and 99.5\% for CNN. The classification errors are lower than the significance level ($\alpha=5\%$). Therefore, we accept the null hypothesis that SGD-based local solutions converge to a similar local solution. Without using the chain normalization rule, the accuracy drops significantly, e.g., 62.9\%, 2.4\%, and 2.8\% for MLP, and 9.7\%, 4.4\%, and 4.0\% for CNN. Without weight normalization, the first layer also obtains a moderate accuracy. The reason is that the first layer only has one column permutation variable and the row of the matrix contains partial discriminative information. Layer-based weight classification/retrieval is not an effective weight similarity measure algorithm due to the lack of the weight normalization. 

\begin{figure*}
  \centering
  \includegraphics[width=1.0\textwidth]{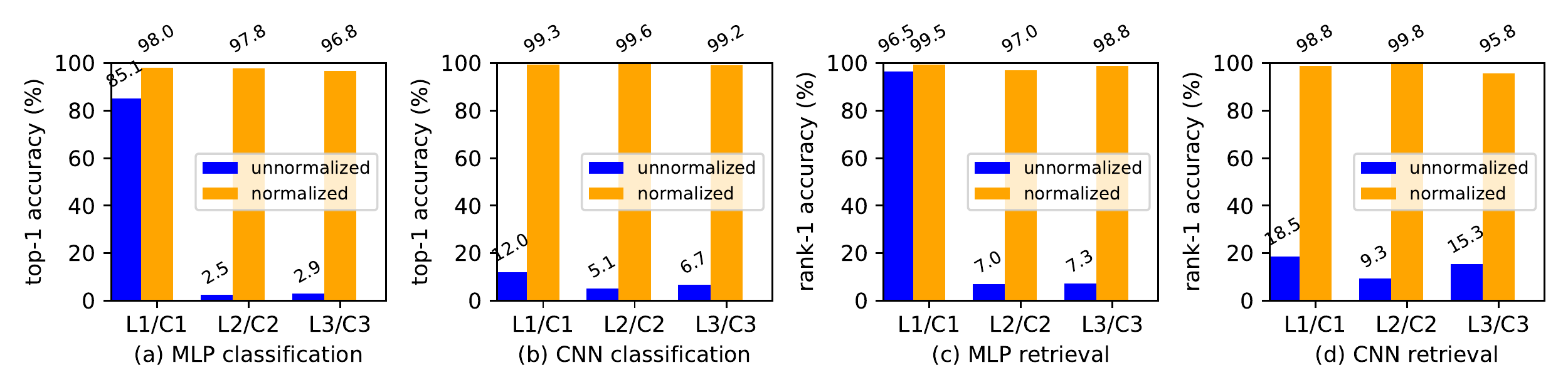}
  \vspace{-22pt}
  \caption{\small{Hypothesis-training-testing on the CIFAR-100 Dataset. L1, L2, and L3 denote the first, second, and third layers and are the x-axis of the unnormalized method. C1, C2, and C3 denote the first, second, and third chains and are the x-axis of the normalized method. In (a) and (b), we adopt the weight classification evaluation protocol for MLP and CNN, respectively. In (c) and (d), we adopt the weight retrieval evaluation protocol for MLP and CNN, respectively.}}\label{fig:cifar100}
  \vspace{-18pt}
\end{figure*}

In weight retrieval, we use the image retrieval metric, i.e., cumulative matching characteristic \cite{Gray2007Evaluating}. As shown in \textbf{Fig. \ref{fig:tiny} (c)} and \textbf{(d)}, our method achieves 99.3\%, 98.8\%, and 98.0\% rank-1 accuracy using MLP. For the weight representation of CNN, the proposed model achieves 98.5\%, 96.3\%, and 97.0\% rank-1 accuracy. The retrieval errors of MLP and CNN are lower than the significance level ($\alpha=5\%$). Therefore, we should accept the hypothesis. Without using the chain normalization rule, the accuracy drops to 85.3\%, 9.5\%, and 5.3\% for MLP, and 15.3\%, 8.8\%, and 11.0\% for CNN. The weight retrieval results show the robustness of the proposed weight representation for unseen weight classes (learning tasks on training weight sets and testing weight sets are non-overlapped), and the high accuracy demonstrates the soundness of the null hypothesis.


\subsection{Hypothesis-training-testing on CIFAR-100}
\label{subsec:cifar100}
The CIFAR-100 dataset \cite{krizhevsky2009learning}, $32\times 32$ in size, has 100 classes.  Each class contains 600 images, including 500 training images and 100 testing images. The null hypothesis and the implementation in this section are similar to that in Section \ref{subsec:tiny}. In both weight classification and retrieval, our method obtains about 98.0\% top-1 accuracy and 98.0\% rank-1 accuracy, as shown in \textbf{Fig. \ref{fig:cifar100}}. The errors are lower than the significance level of $5\%$. Hence, we accept the null hypothesis. The results verify the soundness of the hypothesis and the effectiveness of our proposed weight similarity measure.

\subsection{Hypothesis-training-testing on NameData}
\label{subsection:namadata}
The NameData dataset \cite{paszke2017automatic} contains a few thousand surnames from 18 languages of origin. It is used to train a character-level RNN that can predict from which language a name comes based on the spelling. The hypothesis is similar to Section \ref{subsec:tiny}, but concerns the Recurrent Neural Network (RNN). We train an RNN with two GRU cells (GRU-2) \cite{ChungGCB14} on NameData. A fully connected layer is added after the GRU module for classification.

\begin{wraptable}{r}{8cm}
\vspace{-17pt}
\caption{\small{Hypothesis-training-testing on the NameData Dataset using RNN.}}\label{tab:RNN}
\begin{tabular}{c|c|c}
\hline
                     & unnormalized      & normalized       \\ \hline\hline
Weight classification & 17.2\% top-1  & 100.0\% top-1   \\ \hline
Weight retrieval      & 21.3\% rank-1 & 95.7\% rank-1 \\ \hline
\end{tabular}
\vspace{-16pt}
\end{wraptable} 
\noindent
We evaluate the weight similarity measure of RNN with the weight classification and retrieval protocols. With the chain normalization rule, the proposed method obtains 100.0\% top-1 accuracy at the weight classification setting and 95.7\% rank-1 at the weight retrieval setting. The errors are lower than the significance level of $5\%$. Therefore, we accept the hypothesis. Without the weight normalization approach, the accuracy is 17.2\% in classification and 21.3\% in retrieval. The results show the soundness of the hypothesis and the effectiveness of our weight similarity measure on RNN.

\begin{figure*}
 \vspace{-46pt}
  \centering
  \includegraphics[width=1.0\textwidth]{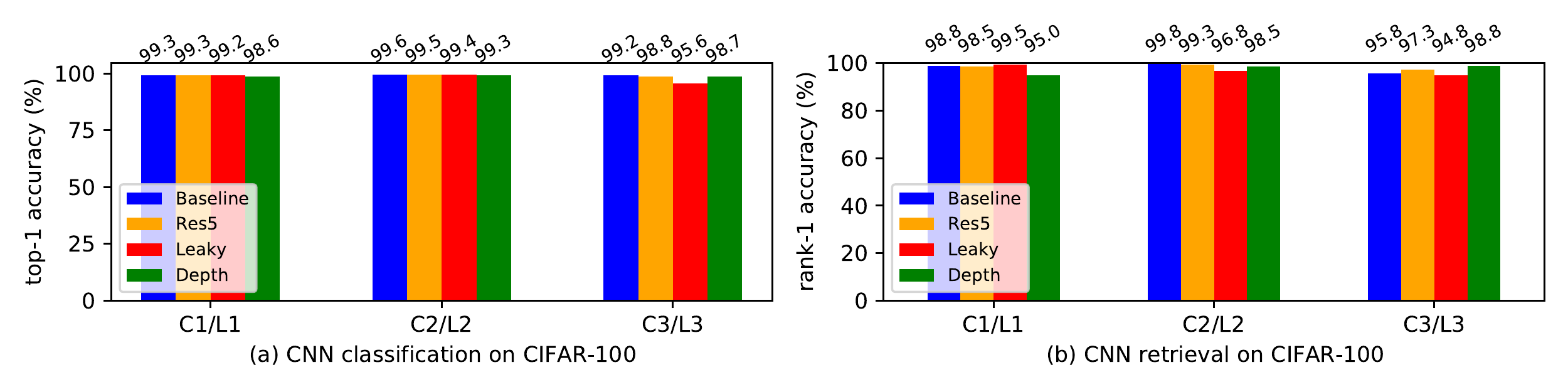}
  \vspace{-25pt}
  \caption{\small{Results of three new networks. C1/L1, C2/L2, and C3/L3 denote the first, second, and third chains/layers.}}\label{fig:other_setting}
  \vspace{-17pt}
\end{figure*}

\begin{figure*}
  \centering
  \includegraphics[width=1.0\textwidth]{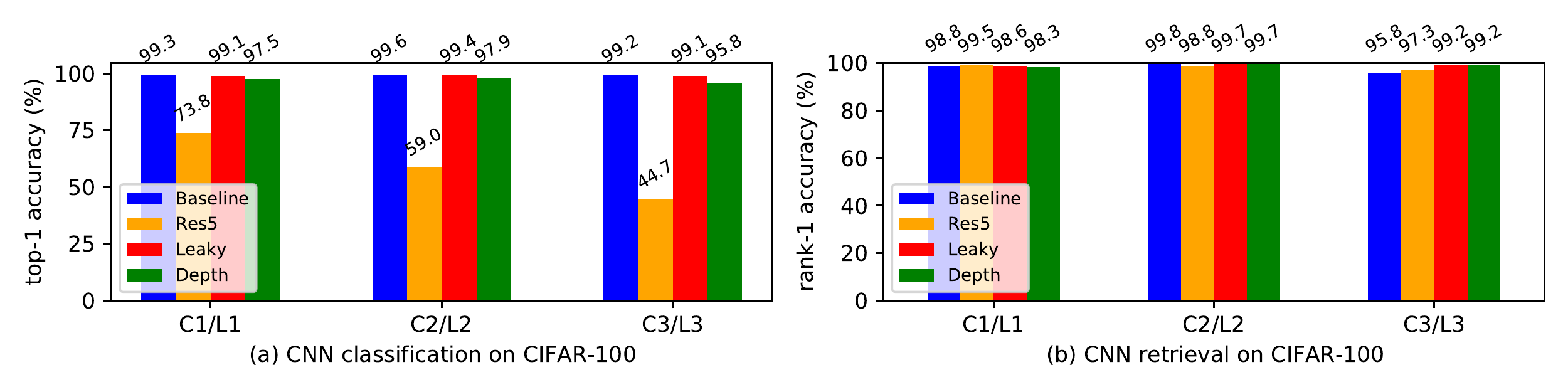}
  \vspace{-22pt}
  \caption{\small{(a) Cross-network weight classification. (b) Cross-network weight retrieval. C1/L1, C2/L2, and C3/L3 denote the first, second, and third chains/layers.} }\label{fig:cross}
  \vspace{-11pt}
\end{figure*}

\subsection{Hypothesis on three new network structures. }
\label{subsec:factors}
We study the hypotheses (Eq. \eqref{eq:h_0}) on three new network structures. We first implement a \textbf{baseline (``Baseline'')} by PlainNet-5 with the ReLU activation function and optimize it by SGD, as discussed in Section \ref{subsec:tiny}. The three new networks are designed upon the Baseline.  
\textbf{1) Depth of Networks (``Depth")}. Depth is implemented by adding a convolutional unit to PlainNet-5, which is referred to as PlainNet-6. \textbf{2) Network structure (``Res5")}. Res5 is designed as a 5-layer residual network (ResNet-5). Note that the weight shape remains the same.  \textbf{3) Activation function (``Leaky")}. Leaky is implemented by replacing all ReLU activation functions with leaky ReLU activation functions. 

Experiments are conducted on the CIFAR-100 dataset. We perform hypothesis-training-testing and study whether the hypothesis under each new network still holds. In \textbf{Fig. \ref{fig:other_setting} (a)} and  \textbf{(b)}, we obtain: \textbf{First}, in both weight classification and retrieval, using the residual network structure, using leaky ReLU activation function, or adding one more layer can still obtain high accuracy. Therefore, the hypothesis still holds. 


\subsection{Hypothesis on cross-network validation}
\textbf{Cross-network weight classification} In this experiment, we study whether changing a factor leads to different weight structures with cross-network classification. Specifically, we train a weight classifier with weights of Net A (Baseline) and use it to predict the weight labels of Net B (Depth, Res5, or Leaky) on the weight set of CIFAR-100 as described in Section \ref{subsec:cifar100}. Note that the weight shape sizes of two nets A and B are the same. As shown in \textbf{Fig. \ref{fig:cross} (a)}, we observe that \textbf{1} when using the weight representation of PlainNet-5 as that of ResNet-5, the accuracy drops; that is, the weight structure of ResNet-5 is greatly changed relative to PlainNet-5; \textbf{2} replacing ReLU by Leaky ReLU or adding one layer to PlainNet-5 nearly preserves the weight structure of the local solutions because the accuracy is still high in the weight classification. 



\textbf{Cross-network weight retrieval} We train the weight classifiers using the local solutions generated under Net A (Baseline). We then use the trained weight classifiers to extract the weight representations generated under Net B (Depth, Res5, or Leaky). If the weight representation trained under Net A is also appropriate for Net B, the weight representation is robust against the cross-condition representation. \textbf{Fig.\ref{fig:cross} (b)} shows that the three factors (i.e., Res5, Leaky, and Depth) do not affect the hypothesis; therefore the weight representation is robust against these three changes. 

\vspace{-6pt}
\section{Conclusion and Discussion}
\label{sec:conc}

\vspace{-6pt}
In this paper, we present a weight similarity measure approach that can measure the weight similarity of non-linear neural networks. To achieve weight similarity measure, we introduce a novel chain normalization rule to normalize the weights of neural networks. Besides, we develop a hypothesis-training-testing statistical inference approach to study the equivalence hypothesis of local solutions on MLP, CNN, and RNN. The experiments on three datasets demonstrate the weight similarity between SGD-based local solutions of identical neural networks in a metric space. Finally, we use weight similarity of neural networks to uncover the relationship of local solutions under different conditions. 


Understanding weight representations of DNNs rather than data representations of DNNs have various compelling properties and applications as follows (but they are all beyond the scope of this paper).

\vspace{-3pt}
\textbf{AutoML.} Existing AutoML methods, represented by neural architecture search and hyperparameter optimization, have critical inefficiency problems, mainly due to the trustworthiness of model ranking. Weight representations may provide a new solution. By studying the similarity of the neural weights, we demonstrate that neural weights are discriminative representations. This inspires us to establish a direct mapping of neural weights to model accuracy. Once someone has established such a mapping, he (she) may be able to directly maximize the model accuracy (e.g., using gradient ascent) to obtain the final neural weights, free of the huge computational complexity of evaluating model performance with training and test data.

\vspace{-3pt}
\textbf{Scene recognition/matching.} Images, videos, audio, 3D scenes, etc., can all be represented by implicit neural representations \cite{dupont2022data}, which is very promising. For example, both the well-known NeRF \cite{mildenhall2020nerf} and DeepSDF \cite{park2019deepsdf} represent a 3D scene as neural network weights. Therefore, our research on the similarity of neural weights can be directly applied to the 3D scene recognition/matching problem because a 3D scene recognition/matching problem reduces to classifying/matching of neural weights.

\vspace{-3pt}
\textbf{Multimodality.} Studying data association between different modalities (e.g., images \& languages, images \& videos) is standard practice, but it is challenging. When various tasks are encoded into neural weight representations, we can obtain relevant knowledge of multimodal data by examining the similarity of weight representations of neural networks.

\vspace{-3pt}
\textbf{Data privacy and security.} Using our study of the similarity of the weights of neural networks may inspire research on data privacy security. Without accessing the data, we can obtain the type of task by matching and identifying the weights of the neural network.

\vspace{-3pt}
\textbf{Functional analysis.} As we have shown that neural weights are discriminative representations, we can represent a dataset as some neural weights. Taking these neural weights as data and feeding them into a new neural network, we can further represent them with other new neural weights. Through repeated loops, we establish the relationship between representation learning and functional analysis.

\vspace{-3pt}
\textbf{Task pre-training.} Different tasks prefer different initialization weights. Using our research on the similarity of neural weights, we can search for initialized weights for a new task. First, we use a natural language model to extract the task embedding from its descriptions. Then, we use this embedding to search for similar pre-training tasks and use its pre-training weights as initialization.

\vspace{-6pt}
\section*{Limitations and broader impact}

\vspace{-6pt}
\textbf{Limitations} In this paper, we propose a weight representation to describe the weights of neural networks. We do not consider the optimization process of non-convex neural networks. Our method provides a statistical explanation of trained models and enables us to better understand equivalent local solutions in a metric space. It is not a study of non-convex optimization theory.

\vspace{-3pt}
\textbf{Broader impact} The proposed hypothesis-training-testing provides a new paradigm for general validation problems, which can be regarded as a "learning-to-explain" way. The proposed weight similarity measure makes a step to open black-box neural networks in a metric space. The method mainly focuses on the statistical explanation of neural networks. We believe there are no potential negative societal impacts.


{\small
\bibliographystyle{plain}
\bibliography{reference}
}

\appendix
\section{Algorithm}
\label{sec:algor}
We summarize an overview of the proposed algorithm for the weight similarity measure, as shown in \textbf{Algorithm} \ref{alg:alg}. We first make a null hypothesis $\mathcal{H}_0$ that the local solutions of an identical neural network are equivalent. To achieve this, we construct a set of learning tasks for weight similarity learning. For each learning task, we generate some weights and assign labels based on the null hypothesis. To transform the weights into ``common data," we propose a chain normalization operator for weight normalization. We then train $L$ weight classifiers for $L$ types of chains. Finally, we adopt the weight classification protocol or the weight retrieval protocol for testing. If the test errors of $L$ chains fall below a pre-defined significance level of $\alpha$, we accept the null hypothesis $\mathcal{H}_0$, and vice versa.

\IncMargin{0.0em}
\begin{algorithm}
\caption{Weight Similarity Measure via Hypothesis-Training-Testing}
\label{alg:alg}
    \SetKwData{Left}{left}
    \SetKwData{Up}{up}
    \SetKwFunction{FindCompress}{FindCompress}
    \SetKwInOut{Input}{input}
    \SetKwInOut{Output}{output}
\Indm\Indmm

\Indp\Indpp
Make a null hypothesis $\mathcal{H}_0$: The local solutions of an identical neural network are equivalent;\\
Construct a set of learning tasks $\Sigma=\{\mathcal{T}_1,\mathcal{T}_2,...,\mathcal{T}_M\}$;\\
\For{ $j=1:M$ }{Generate $N_j$ weights for $\mathcal{T}_j$, which is denoted as $\{W^{\mathcal{T}_j,i}\}_{i=1}^{N_j}$ and assigned a weight label $\mathcal{Y}_{\mathcal{T}_j}$ based on $\mathcal{H}_0$;\\}
\For{ $l=1:L$ }{
Normalize the weights by the weight normalization operator $\phi$ according to Eq. (7);\\ 
Train a weight classifier $F_l(\cdot;\Theta_l)$ by Eq. (8);\\
Make a test according to the weight classification protocol or the weight retrieval protocol, and returns a test error $e_l$; \\}
For any $l$, if the test error $e_l$ is lower than a pre-defined significance level $\alpha$, we accept the null hypothesis $\mathcal{H}_0$, and vice versa;\\
\end{algorithm}

\section{A Proof of Chain Normalization Rule}
\label{appendix:proof}
We begin by considering the weight $W_1\in \mathbb{R}^{n_0\times n_1}$ because $W_1$ is only affected by $W_2$. If we  permute the $i$-th and $j$-th neurons of $h_1$ and wish to keep the output unchanged, we have to permute the $i$-th and $j$-th columns of $W_1$ and $i$-th and the $j$-th rows of $W_2$ correspondingly, as illustrated in \textbf{Fig. 1}. Let $Q_1\in \mathbb{R}^{n_1\times n_1}$ be a column permutation matrix (i.e., a matrix formed by permuting the columns of the identity matrix $I\in \mathbb{R}^{n_1\times n_1}$) and $W_1^{*}$ be the normalized form of $W_1$. The weight normalization of $W_1$ is given by


\begin{equation}
\label{eq:tranform}
W_1Q_1=W_1^{*}.
\end{equation}


In Eq. \eqref{eq:tranform}, we cannot directly obtain $W_1^{*}$, because both $W_1^{*}$ and $Q_1$ are unknown. In fact, $W_1$ is not fixed in different runs, so $Q_1$ is not fixed either. To find a weight normalization operator, we have to  eliminate the permutation factor $Q_1$. In Eq. \eqref{eq:tranform}, both sides multiply its transpose. We  obtain
\begin{equation}
\label{eq:tranform2}
(W_1Q_1)(W_1Q_1)^{T}=W_1^{*}W_1^{*T}.
\end{equation}
Because $Q_1$ is the permutation of the identity matrix $I$,  $Q_1$ is a normalized orthogonal matrix. Hence, $Q_1Q_1^{T}=I^{n_1\times n_1}$. Eq. \eqref{eq:tranform2} can be simplified as
\begin{equation}
\label{eq:tranform2_2_}
W_1W_1^{T}=W_1^{*}W_1^{*T}.
\end{equation}
Because $W_1^{*}W_1^{*T}$ is a normalized value and is fixed, $W_1W_1^{T}$ is invariant to the random permutation $Q_1$.

We then consider $W_2$. $W_2$ is affected by the column permutation of $W_1$ and the row permutation of $W_3$. Given any $W_2$, suppose that a row permutation matrix $P_2\in \mathbb{R}^{n_1\times n_1}$ and a column permutation matrix $Q_2\in \mathbb{R}^{n_2\times n_2}$ exist such that $W_2$ can be transformed into $W_2^{*}$. We have
\begin{equation}
\label{eq:tranform3}
P_2W_2Q_2=W_2^{*}.
\end{equation}
In Eq. \eqref{eq:tranform3}, it is difficult to eliminate both $P_2$ and $Q_2$. Considering Eqs. \eqref{eq:tranform} and \eqref{eq:tranform3}, both sides are multiplied, respectively. We have
\begin{equation}
\label{eq:tranform4}
(W_1Q_1)(P_2W_2Q_2)=(W_1^{*})(W_2^{*}).
\end{equation}
Because the normalization of $W_2$ is jointly affected by $W_1$, we have $Q_1=P_2^{T}$, as illustrated in \textbf{Fig. 1}. $Q_2$ and $P_2$ are normalized orthogonal matrixes. Hence, $Q_1P_2=P_2^TP_2=I^{n_1\times n_1}, Q_2Q_2^{T}=I^{n_2\times n_2}$.  We simplify Eq. \eqref{eq:tranform} as
\begin{equation}
\label{eq:tranform5}
W_1W_2Q_2=W_1^{*}W_2^{*}.
\end{equation}
Similar to Eq. \eqref{eq:tranform2}, we eliminate the permutation factor $Q_2$ by
\begin{equation}
\label{eq:tranform5_}
(W_1W_2Q_2)(W_1W_2Q_2)^T=(W_1^{*}W_2^{*})(W_1^{*}W_2^{*})^T.
\end{equation}
Hence, we obtain
\begin{equation}
\label{eq:tranform6}
W_1W_2W_2^TW_1^T=W_1^{*}W_2^{*}W_2^{T*}W_1^{T*}.
\end{equation}
In this way, we can easily generalize Eq. \eqref{eq:tranform6} to $W_l$, which is given by 
\begin{equation}
\label{eq:tranform7}
W_1W_2...W_lW_l^T...W_2^TW_1^T=W_1^{*}W_2^{*}...W_l^{*}W_l^{*T}...W_2^{T*}W_1^{T*}.
\end{equation}
It is observed that the left side of Eq. \eqref{eq:tranform7} is independent of permutation factors because the right side comprises a sequence of normalized weights and is thus fixed. We term Eq. \eqref{eq:tranform7} as the \emph{chain normalization rule} for the weight of neural networks. Here, a chain is defined as a sequence of layers of a neural network that begins with the first layer. This implies that the $l$-th chain is from the 1-st layer to the $l$-th layer. The chain normalization rule suggests that we can normalize the weights of neural networks based on chains instead of layers. Therefore, the weight normalization operator $\phi(\cdot)$ is given by
\begin{equation}
\label{eq:tranform8_}
\phi(W_1,W_2,...,W_l)=W_1W_2...W_lW_l^T...W_2^TW_1^T,
\end{equation}
where $\phi(\cdot)$ takes a chain (from $W_1$ to $W_l$) as the input and produces a matrix as a normalized weight. In this way, we transform the layer-based weights $W_1$, $W_2$, ... , $W_L$ into the chain-based weights $\phi(W_1)$, $\phi(W_1, W_2)$, ... , $\phi(W_1, W_2,...,W_L)$, or briefly denoted as $\phi^1(W)$, $\phi^2(W)$, ... , $\phi^L(W)$ for simplification. This implies that a weight of a neural network with $L$ layers is split into $L$ chains (not layers) for normalization, respectively. We cannot normalize a neural network layer by layer because we cannot eliminate both permutation factors $P_l$ and $Q_l$ simultaneously.

\textcolor{black}{\textbf{Remark.} Different from MLP, $W_l$ of CNN is a 4-dim tensor. We need to reshape tensors for CNN. Suppose all of the kernel sizes are $h$ and $w$. Let $W_{1}\in \mathbb{R}^{{n_0}\times {n_1} \times h \times w}$ and $W_{2}\in \mathbb{R}^{{n_1}\times {n_2} \times h \times w}$, we first reshape $W_{1}$ into $W^{'}_{1} \in \mathbb{R}^{{n_0}hw\times {n_1}}$ and reshape $W_{2}$ into $W^{'}_{2} \in \mathbb{R}^{{n_1}\times {n_2}hw}$. We then compute $W^{'}_{1} W^{'}_{2} \in \mathbb{R}^{{n_0}hw\times {n_2}hw}$, which is then reshaped into $\mathbb{R}^{{n_0}(hw)^2\times {n_2}}$ and multiplies $W_3$. In this way, we obtain $W_1^{'}W_2^{'}...W_l^{'} \in \mathbb{R}^{{n_0}(hw)^{l-1}\times {n_l}}$. $\phi(W_1^{'},W_2^{'},...,W_l^{'})\in \mathbb{R}^{{n_0}(hw)^{l-1}\times {n_0}(hw)^{l-1}}$.} 

\section{Networks and Implementation} The PlainNet-5 network consists of four convolutional units and one fully-connected layer. Each convolutional unit contains a convolutional layer with a kernel size of $3\times 3$, a ReLU activation layer, a BatchNorm layer, and a pooling layer. The MLP-4 consists of four fully connected layers (each of which is followed by a ReLU function); the first three layers have 500 hidden neurons, and the last layer has $4$ neurons.  The RNN in Section 4.3 contains two GRU cells (GRU-2). We use one GeForce RTX2080TI in our experiments.

\section{Weight similarity measure on deeper networks}
\textcolor{black}{To study the weight similarity measure on deeper neural networks, we implement an MLP-based version of ResNet, i.e., ResNet-MLP-12. The main structure of ResNet-MLP-12 is similar to ResNet-18. ResNet-MLP-12 contains 6 residual building blocks, each of which has two MLP layers (Linear+ReLU+BN). Note that each task only contains 2 classes and each class contains 500 training images in our CIFAR-100 setting. We do not implement 18 layers because training a ResNet-MLP-18 with 1,000 images ($32\times32$) leads to over-fitting. Therefore, we use ResNet-MLP-12. Since we focus on studying the weight similarity of neural networks, ResNet-MLP-12 is a 12-layer deep neural network and is thus sufficient to validate our hypothesis. The experiment setting is similar to Section 4.2. We use the weight classification evaluation protocol. As shown in \textbf{Table \ref{tab:ResNet-MLP-12}}, the errors of different chains are lower than a pre-defined significance level of $\alpha=5\%$, we accept the null hypothesis. This indicates that our proposed method can be extended to deeper neural networks.}


\begin{table}[!h]
\caption{Weight similarity measure on CIFAR-100 using ResNet-MLP-12.}\label{tab:ResNet-MLP-12}
\begin{center}
\begin{tabular}{c|c|c|c|c|c}
\hline
\hline
Chain number & 3 &   5   & 7 & 9   & 11     \\ \hline
top-1 accuracy    & 100.0\%  & 100.0\%  & 100.0\% & 99.7\%  & 100.0\% \\ \hline\hline
\end{tabular}
\end{center}
\end{table}

\section{Detailed Implementation on CIFAR-100}
\label{subsec:cifar100_}
The CIFAR-100 dataset \cite{krizhevsky2009learning}, $32\times 32$ in size, has 100 classes.  Each class contains 600 images including 500 training images and 100 testing images. The null hypothesis in this section is similar to that in Section 4.1.
\subsection{Training}
We train PlainNet-5 and MLP-4 to form two weight sets on CIFAR-100, respectively. The 100 classes of CIFAR-100 are divided into 50 groups as 50 data subsets. Each data subset has 2 image classes and represents a learning task. There are 50 learning tasks in total, whose weight labels range from 0 to 49.  For each learning task, we repeat the training procedure 100 times to obtain 100 local weight. Finally, we obtain 5,000 weights for MLP and CNN, respectively.

The implementation of CIFAR-100 is similar to that of tiny ImageNet. The structure of PlainNet-5 and MLP-4 differs slightly from previous ones because each data subset of CIFAR-100 contains 2 classes. We modify the dimension of the last fully connected layer from 4 to 2.
\subsection{Testing}
In both local weight classification and retrieval, our method obtains about 98.0\% top-1 accuracy and 98.0\% rank-1 accuracy, as shown in \textbf{Fig. 3}. The errors are lower than the significance level of $5\%$. Therefore, we accept the null hypothesis. These results once again demonstrate the soundness of the hypothesis and the effectiveness of our proposed weight similarity measure.
\begin{table}[!h]
\caption{Hypothesis-training-testing on the NameData Dataset using RNN.}\label{tab:RNN_}
\begin{center}
\begin{tabular}{c|c|c}
\hline
\hline
                     & unnormalized      & normalized       \\ \hline
local weight classification & 17.2\% top-1  & 100.0\% top-1   \\ \hline
local weight retrieval      & 21.3\% rank-1 & 95.7\% rank-1 \\ \hline\hline
\end{tabular}
\end{center}
\end{table}

\section{Detailed Implementation on NameData}
\label{subsection:namadata_}
The NameData dataset \cite{paszke2017automatic} contains a few thousand surnames from 18 languages of origin. It is used to train a character-level RNN that can predict from which language a name comes based on the spelling. The hypothesis is similar to Section 4.1 but concerns the Recurrent Neural Network (RNN).

\subsection{Training}
We train an RNN with two GRU cells (GRU-2) \cite{ChungGCB14} on NameData. A fully connected layer is added after the GRU module for classification. We divide 18 classes into 9 groups as 9 data subsets. Each data subset contains 2 classes and represents a learning task. Therefore, there are 9 learning tasks in total, whose weight labels range from 0 to 8. We repeat the training procedure 100 times to obtain 100 weights for each class. Finally, we generate 900 weights.

When generating the weight set, we use SGD with a mini-batch size of 1. The learning rate starts from 0.1 and is divided by 10 after 7,000 batches. We train the RNN for 10,000 batches in total. When training the weight classifier, we use SGD with a mini-batch size 1. The learning rate is 0.001 and is divided by 10 after 6 epochs. We train the weight classifier for 10 epochs.


\subsection{Testing}
We evaluate the weight similarity measure of RNN with the weight classification and retrieval protocols as discussed in Section 4.1. With the chain normalization rule, the proposed method obtains 100.0\% top-1 accuracy at the weight classification setting and 95.7\% rank-1 at the weight retrieval setting. The errors are lower than the significance level of $5\%$. Therefore, we accept the hypothesis. Without the normalization approach, the accuracy is 17.2\% in classification and 21.3\% in retrieval. These results demonstrate the soundness of the hypothesis and the effectiveness of our proposed weight similarity measure on RNN.

\end{document}